\documentclass{book}    
\usepackage{piers}  
\usepackage{booktabs}
\usepackage{float}
\usepackage{subfigure}
\usepackage{hyperref}

\begin{document}

\title{Hyperspectral Image Analysis with Subspace Learning-based
One-Class Classification}
\maketitle

\author      {F. M. Lastname}
\affiliation {University}
\address     {}
\city        {Boston}
\postalcode  {}
\country     {USA}
\phone       {345566}    
\fax         {233445}    
\email       {email@email.com}  
\misc        { }  
\nomakeauthor

\author      {F. M. Lastname}
\affiliation {University}
\address     {}
\city        {Boston}
\postalcode  {}
\country     {USA}
\phone       {345566}    
\fax         {233445}    
\email       {email@email.com}  
\misc        { }  
\nomakeauthor

\begin{authors}

{\bf Sertac Kilickaya}$^{1}$, {\bf Mete Ahishali}$^{2}$, {\bf Fahad Sohrab}$^{2}$, {\bf Turker Ince}$^{1}$, {\bf and Moncef Gabbouj}$^{2}$\\
\medskip
$^{1}$Department of Electrical and Electronics Engineering, Izmir University of Economics, Izmir, Turkey\\
$^{2}$Department of Computing Sciences, Tampere University, Tampere, Finland

\end{authors}

\begin{paper}

\begin{piersabstract}
Hyperspectral image (HSI) classification is an important task in many applications, such as environmental monitoring, medical imaging, and land use/land cover (LULC) classification. Due to the significant amount of spectral information from recent HSI sensors, analyzing the acquired images is challenging using traditional Machine Learning (ML) methods. As the number of frequency bands increases, the required number of training samples increases exponentially to achieve a reasonable classification accuracy, also known as the curse of dimensionality. Therefore, separate band selection or dimensionality reduction techniques are often applied before performing any classification task over HSI data. In this study, we investigate recently proposed subspace learning methods for one-class classification (OCC). These methods map high-dimensional data to a lower-dimensional feature space that is optimized for one-class classification. In this way, there is no separate dimensionality reduction or feature selection procedure needed in the proposed classification framework. Moreover, one-class classifiers have the ability to learn a data description from the category of a single class only. Considering the imbalanced labels of the LULC classification problem and rich spectral information (high number of dimensions), the proposed classification approach is well-suited for HSI data. Overall, this is a pioneer study focusing on subspace learning-based one-class classification for HSI data. We analyze the performance of the proposed subspace learning one-class classifiers in the proposed pipeline. Our experiments validate that the proposed approach helps tackle the curse of dimensionality along with the imbalanced nature of HSI data.
\end{piersabstract}

\psection{Introduction}
 Hyperspectral imaging has become an essential tool in various fields, such as remote sensing, geology, and agriculture \cite{SpaBS1, SpaBS2}. Due to the high-dimensional nature of hyperspectral data, which can have hundreds of frequency bands, traditional machine learning algorithms face challenges in efficiently and effectively classifying this data. The curse of dimensionality is a challenge in which the required number of training samples increases exponentially as the number of frequency bands increases, making it difficult to achieve reasonable classification accuracies. Furthermore, in the case of HSI data, the labels can be highly imbalanced, which means that there can be a large number of samples from one class and a much smaller number from other classes. This can again make traditional ML methods ineffective in generalization as they may overfit the majority class and perform poorly on the minority classes. One-class classifiers (OCC), on the other hand, are very useful in cases where the training data available is from a single class only. Such classifiers can be used to train a model with features that are characteristic of the target class only.
 
\par
Spectral bands in hyperspectral images are often highly correlated. Therefore, band selection/reduction methods \cite{srl-soa, SpaBS3, ISSC1} are often employed as a preprocessing step for hyperspectral image analysis to improve classification accuracy and reduce computational costs. Two common techniques used for this task are feature extraction and feature selection. Feature extraction transforms the original high-dimensional space of raw hyperspectral images to a lower-dimensional feature space using algorithms like PCA \cite{agarwal2007efficient}, or ICA \cite{du2008similarity}. However, a drawback of feature extraction methods is that the transformed feature space may no longer contain spectral information and can lead to the loss of interpretability in hyperspectral data. The other approach is called feature or band selection, and it aims to identify the most informative bands for classification. Different methods have been proposed for band selection, and they can be categorized into supervised \cite{yang2010efficient}, unsupervised \cite{guo2006band} \cite{geng2014fast}, and semi-supervised \cite{bai2015semisupervised} approaches. While band selection can improve the performance of hyperspectral image analysis, there are also some potential drawbacks to consider. First of all, the selected subset of bands may not be optimal for all classification tasks and could result in reduced accuracy for certain applications. Moreover, the computational complexity and potential bias introduced by the band selection techniques may limit their applications as a preprocessing step in classification. For example, supervised band selection methods can be biased toward the classes that are overrepresented in the training data. In contrast, unsupervised methods may miss important information that is not apparent in the data distribution. Finally, there is a risk of overfitting when using dimensionality reduction techniques, where the model may fit the existing noise instead of the underlying patterns in the data. This can lead to poor generalization performance on new data.
\par
To mitigate these challenges, in this study, we leverage Subspace Support Vector Data Description (S-SVDD) for the one-class classification of hyperspectral images. The S-SVDD method is a hybrid technique that combines the advantages of subspace learning and support vector data description (SVDD) \cite{tax2004support}. S-SVDD employs a subspace learning technique to project the data onto a lower-dimensional feature space that optimizes the one-class classification task; hence there is no need for separate dimensionality reduction or feature selection procedure in the proposed classification framework. It aims to learn the boundary of the target data distribution in the subspace by a hyper-sphere that encloses the positive samples and maximizes the margin between the sphere and the negative samples. This boundary defines the one-class region in the low-dimensional feature space, which can be used to classify new data points. The S-SVDD method has been shown to be effective in handling the curse of dimensionality and imbalanced data while also avoiding the risk of overfitting \cite{sohrab2018subspace}, and this research can be considered a pioneer study on subspace learning-based one-class classification for HSI data. The inference pipeline for S-SVDD approach with linear mapping is illustrated in Figure 1. The performance of the subspace learning-based one-class classifier is analyzed in the proposed pipeline on two benchmark HSI datasets: Salinas-A \cite{salinas} and Indian Pines \cite{indian}. Our experiments demonstrate the usage of different regularization terms proposed in the S-SVDD method \cite{sohrab2018subspace}, and validate the usage of the proposed approach to tackle the curse of dimensionality along with the imbalanced nature of HSI data.

\begin{figure}[htp]
    \centering
    \includegraphics[width=17cm]{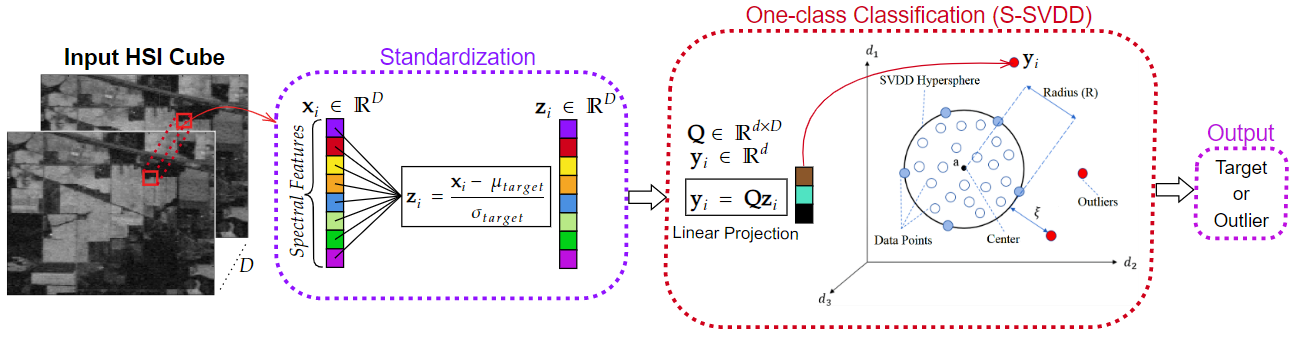}
    \caption{One-class Classification of HSI Images using S-SVDD Approach} 
    \label{fig:ssvdd}
\end{figure}

\psection{Methodology}
One-class classification focuses on creating a representative model for a specific class of interest, commonly referred to as the target or positive class \cite{sohrab2023graph}. This model is developed using data exclusively from the target class. When making predictions, the model is used to determine whether new or unseen samples are part of the target class or if they are outliers. While traditional one-class classifiers are capable of generating successful class models using only a limited number of samples, the computational complexity associated with the training phase makes it infeasible for large or high-dimensional datasets. 

\par
In one-class learning, we have samples from positive class only, and the task is to find a boundary that can detect outliers lying outside of the converged boundary. In SVDD, in particular, the algorithm tries to find minimal circumscribing hyperball that comprises only the positive observations in high-dimensional space. Given the positive data representations in the high-dimensional feature space $\mathbb{R}^{D}$, we need to determine the center of the class $ \mathbf{a} \in \mathbb{R}^{D} $ and the radius $R$ of the hypersphere, by minimizing the following:
\begin{equation*} 
F(R, \mathbf{a})=R^{2}+C\sum_{i=1}^{N}\xi_{i},
\tag{1} 
\end{equation*}
such that the following conditions are 	fulfilled, i.e.:
\begin{align*} &\Vert\mathbf{x}_{i}-\mathbf{a}\Vert_{2}^{2} \leq R^{2}+\xi_{i}, \quad i=1, \ldots, N \tag{2}\\ &\qquad\qquad\quad\xi_{i}\ \geq 0,\quad\ i=1,\ \ldots,\ N. \tag{3} \end{align*}
where the parameter $C>0$ is a regularization parameter that controls the trade-off between the volume of the hypersphere and the training error caused by allowing outliers in the class description, and $\xi_{i}$ is the set of slack variables. When $C$ is increased, we have a tighter hypersphere, and training error is decreased. However, $C$ should be chosen very carefully because one may end up overfitting the training data if $C$ is very large. Slack variables are introduced in these equations as they allow us to handle the possibility of outliers in the training data.

The S-SVDD method, on the other hand, attempts to identify a $d$-dimensional feature space ($d<D$), that can optimally represent the class. In the case of linear projection, the aim is to find a matrix $ \mathbf{Q}\in\mathbb{R}^{d \times D} $ such that:
\begin{equation*} 
\mathbf{y}_{i}=\mathbf{Qx}_{i}, \quad i=1, \ldots, N, 
\tag{4} 
\end{equation*}
Now, one can rewrite the constraints for S-SVDD similar to SVDD as:

\begin{align*} &\Vert\mathbf{Qx}_{i}-\mathbf{a}\Vert_{2}^{2} \leq R^{2}+\xi_{i}, \quad i=1, \ldots, N \tag{5}\\ &\qquad\qquad\quad\xi_{i}\ \geq 0,\quad\ i=1,\ \ldots,\ N. \tag{6} \end{align*}

As previously mentioned, the objective is to find a minimal circumscribing hypersphere that comprises the positive observations, but this time in low-dimensional space d; thus, the same objective function in Equation 1 can be used for S-SVDD. The only addition to the original SVDD problem is the matrix $\mathbf{Q}$. To find the optimal parameters, we can follow Lagrange-based optimization steps. We can build the Lagrangian which has both primal ($R, \mathbf{a},\mathbf{Q}, \xi_{i}$) and dual ($\alpha_{i}\ge0, \gamma_{i}\ge0$) variables as:

\begin{align*} L(R,\mathbf{a}, \alpha_{i}, \xi_{i}, \gamma_{i}, \mathbf{Q}) &= R^{2}+C\sum_{i=1}^{N}\xi_{i}-\sum_{i=1}^{N}\gamma_{i}\xi_{i}- \sum_{i=1}^{N}\alpha_{i}\left(\right.R^{2}+\xi_{i}-\mathbf{x}_{i}^{\intercal}\mathbf{Q}^{\intercal}\mathbf{Qx}_{i}+ 2\mathbf{a}^{\intercal}\mathbf{Qx}_{i}-\mathbf{a}^{\intercal}\mathbf{a}). \tag{7} \end{align*}
Now, we should minimize the Lagrangian with respect to primal variables radius $R$, center $a$, slack variables $\xi_{i}$, and projection matrix $\mathbf{Q}$. Differentiating the Lagrangian with respect to primal variables and setting the partial derivatives equal to zero gives:

\begin{align*} &\frac{\partial L}{\partial R}=0 \Rightarrow \displaystyle \sum_{i=1}^{N}\alpha_{i}=1, \tag{8}\\ &\frac{\partial L}{\partial \mathbf{a}}=0 \Rightarrow \mathbf{a}=\sum_{i=1}^{N}\alpha_{i}\mathbf{Qx}_{i}, \tag{9}\\ &\frac{\partial L}{\partial\xi_{i}}=0 \Rightarrow C-\alpha_{i}-\gamma_{i}=0, \tag{10}\\ &\frac{\partial L}{\partial \mathbf{Q}}=0 \Rightarrow \mathbf{Q}=\left(\sum_{i=1}^{N}\alpha_{i}\mathbf{x}_{i}\mathbf{x}_{i}^{\intercal}\right)^{-1}\left(\sum_{i=1}^{N}\alpha_{i}\mathbf{x}_{i}\mathbf{a}^{\intercal}\right). \tag{11} \end{align*}

It is evident from Equations (8) to (11) that the optimization parameters $\alpha_{i}$ and $\mathbf{Q}$ are interconnected; hence they cannot be optimized together. By inserting the optimal values for the primal variables into Equation 7, we can write the Wolfe dual as:
\begin{equation*} L=\sum_{i=1}^{N}\alpha_{i}\mathbf{y}_{i}^{\intercal}\mathbf{y}_{i}-\sum_{i=1}^{N}\sum_{j=1}^{N}\alpha_{i}\mathbf{y}_{i}^{\intercal}\mathbf{y}_{j}\alpha_{j}. \tag{12} \end{equation*}
Wolfe dual needs to be maximized with respect to dual variables $\alpha_{i}$. The support vectors that define the class description are represented by samples $\mathbf{y}_{i}=\mathbf{Qx}_{i}$, where $\alpha_{i}>0$. Samples with $0<\alpha_{i}<C$ have corresponding $\mathbf{y}_{i}$ values on the boundary of the hypersphere, while samples outside the boundary have $\alpha_{i}=C$. For samples inside the boundary, their corresponding $\alpha_{i}$ values are equal to zero. It is worth noting that whether a sample is a support vector or not is impacted by the selection of the data projection matrix $\mathbf{Q}$. Therefore, in the optimization process, we start with a random $\mathbf{Q}$, which is orthogonalized using QR decomposition and row normalized using $l_{2}$ norm. Then, at each iteration, we first project the data to a lower dimensional space d using (4) and calculate $\alpha_{i}$ by maximizing (12). After finding the set of $\alpha_{i}$, an augmented version of the Lagrangian function is used in the optimization process.
\begin{equation*} L=\sum_{i=1}^{N}\alpha_{i}\mathbf{x}_{i}^{\intercal}\mathbf{Q}^{\intercal}\mathbf{Qx}_{i}-\sum_{i=1}^{N}\sum_{j=1}^{N}\alpha_{i}\mathbf{x}_{i}^{\intercal}\mathbf{Q}^{\intercal}\mathbf{Qx}_{j}\alpha_{j}+\beta\psi, \tag{13} \end{equation*}
where $\beta$ is a regularization parameter controlling the importance of the regularization term $\psi$ in the update and $\psi$ enforces variance in the low-dimensional space which is defined as,
\begin{equation*} \psi=tr(\mathbf{QX}\lambda\lambda^{\intercal}\mathbf{X}^{\intercal}\mathbf{Q}^{\intercal}). 
\tag{14} 
\end{equation*}
 where $tr(\cdot)$ is the trace operator.
 
 Depending on the values of $\lambda$, the regularization term $\psi$ can take different forms. When $\lambda$ is set to zero for all samples, the regularization term becomes obsolete. This is referred to as $\psi_0$. If $\lambda$ is set to one for all samples, all training samples contribute equally to the regularization term $\psi$. This means that all samples are used to describe the variance of the class, and this is referred to as $\psi_1$. If $\lambda$ is set to $\alpha_i$ for each sample $i$, the samples on the class boundary as well as the outliers, are used to describe the class variance and regularize the update of the projection matrix. This is referred to as $\psi_2$. Finally, when $\lambda$ is set to $\alpha_i$ for instances corresponding only to the class boundary and zero for non-support vectors, this is referred to as $\psi_3$. We can finally update $\mathbf{Q}$ by using the gradient of $L$ in (13) as,

\begin{equation*}\mathbf{Q}=\mathbf{Q}-\eta\triangle L,
\tag{15}\end{equation*}
where $\eta$ is the learning rate. In this study, we use different regularization terms of S-SVDD and compare their performance for HSI classification tasks. When making predictions on test data, the test sample is mapped to the low-dimensional space $\mathrm{y}_{\ast}\in \mathbb{R}^{d}$, and the distance from the hypersphere center is computed from the following:

\begin{equation*} \Vert \mathbf{y}_{\ast}-\mathbf{a}\Vert_{2}^{2}=\mathbf{y}_{\ast}^{\intercal}\mathbf{y}_{\ast}-2\sum_{i=1}^{N}\alpha_{i}\mathbf{y}_{\ast}^{\intercal}\mathbf{y}_{i}+\sum_{i=1}^{N}\sum_{j=1}^{N}\alpha_{i}\alpha_{j}\mathbf{y}_{i}^{\intercal}\mathbf{y}_{j}, \tag{16} \end{equation*}
where $\mathbf{y}_{\ast}$ is labelled as the target class if $\Vert \mathbf{y}_{\ast}-\mathbf{a}\Vert_{2}^{2}\leq R^{2}$, otherwise, as outlier. 

In this study, we conducted a comparison between linear and non-linear S-SVDD. To enable non-linear S-SVDD, we leveraged the Non-Linear Projection Trick (NPT) \cite{kwak2013nonlinear} technique. NPT transforms the data at the beginning of the process and allows the subsequent optimization to be performed using the linear method. To achieve this, the kernel matrix is first computed using a kernel function, such as the Radial Basis Function (RBF) kernel, and then centralized. The resulting centralized kernel matrix is then decomposed through eigendecomposition to obtain the reduced dimensional kernel space \cite{sohrab2020ellipsoidal}. This reduced dimensional kernel space is utilized instead of the training data matrix within the original linear method, leading to a non-linear transformation. This approach offers an alternative to the kernel trick. 

\psection{Experimental Evaluation}

\psubsection{Experimental Setup}
We conducted experiments on two widely used hyperspectral image (HSI) datasets, Salinas-A \cite{salinas} and Indian Pines \cite{indian}. The Indian Pines dataset was collected by the Airborne Visible/Infrared Imaging Spectrometer (AVIRIS) sensor over an agricultural area in Indiana, USA. It consists of $145 \times 145$ pixels and 224 spectral bands, covering a spectral range of 0.4-2.5 µm. The number of bands was reduced to 200 by removing the bands covering the region of water absorption, and the number of classes in this dataset is 16. The Indian Pines dataset exhibits a significant class imbalance, with the number of samples per class ranging from 20 to 2455. Due to this imbalance, it is suitable for developing one-class classifiers.

The Salinas dataset was also collected by AVIRIS sensor over an agricultural area in Salinas Valley, California, USA. It consists of $512 \times 217$ pixels and 224 spectral bands. Similar to the Indian Pines data, 20 water absorption bands are discarded. On the other hand, Salinas-A data, which consists of $86 \times 83$ pixels, is a small subscene of the Salinas image, and it includes 6 classes.

\begin{table}[]
\centering
\caption{Number of samples for each class in Salinas-A and Indian Pines dataset are given in (a) and (b), respectively.}
\label{tab:samples}
\begin{subtable}{(a)}
\begin{tabular}{ccc}
\toprule
     & \textbf{Class}               & \textbf{Samples} \\ \midrule
1           & Brocoli Green Weeds     & 391              \\
2           & Corn Senesced Green Weeds & 1343             \\
3           & Lettuce Romaine 4 wk.        & 616              \\
4           & Lettuce Romaine 5 wk.        & 1525             \\
5           & Lettuce Romaine 6 wk.        & 674              \\
6           & Lettuce Romaine 7 wk.        & 799              \\ \bottomrule\end{tabular}
\end{subtable}
\begin{subtable}{(b)}
\begin{tabular}{ccc}
\toprule
     & \textbf{Class}               & \textbf{Samples} \\ \midrule
1           & Alfalfa                      & 46               \\
2           & Corn (notill)                  & 1428             \\
3           & Corn (mintill)                 & 830              \\
4           & Corn                         & 237              \\
5           & Grass (pasture)                & 483              \\
6           & Grass (trees)                  & 730              \\
7           & Grass Pasture Mowed          & 28               \\
8           & Hay (windrowed)                & 478              \\
9           & Oats                         & 20               \\
10          & Soybean (notill)               & 972              \\
11          & Soybean (mintill)              & 2455             \\
12          & Soybean (clean)                & 593              \\
13          & Wheat                        & 205              \\
14          & Woods                        & 1265             \\
15          & Buildings Grass Trees Drives & 386              \\
16          & Stone Steel Towers           & 93               \\ \bottomrule
\end{tabular}
\end{subtable}
\end{table}

To evaluate the performance of the proposed method on these datasets, we followed the following experimental setup. Both scenes are vectorized since the proposed approach is evaluated pixel-wise. For each dataset, we select $30\%$ and $70\%$ percent for training and testing, respectively. These sets are formed by keeping the proportions of samples among different classes equal to the original dataset before splitting. As it is presented in Table \ref{tab:samples}, the datasets are highly imbalanced and for some classes, available annotated data is very limited. The data is then normalized by subtracting the mean and dividing it by the standard deviation which are both computed using only the target class samples from the training set. During the training of the proposed approach, only target class samples from the training set are used. We developed various S-SVDD classifiers with different regularizations $\psi_{0-3}$ and mappings, each dedicated to a particular target class, by training separate models on feature vectors extracted from the training HSI images of the respective target class. In addition, both linear and nonlinear variants of the proposed approach are utilized in our experiments. For the non-linear approach, we specifically employed the NPT with RBF kernel; i.e. $\mathbf{K}_{ij}=\exp\left(\frac{-\Vert \mathbf{x}_{i}-\mathbf{x}_{j}\Vert_{2}^{2}}{2\sigma^{2}}\right)$, where $\sigma$ is an additional hyper-parameter to scale the Euclidean distance between two feature vectors $\mathbf{x}_{i}$ and $\mathbf{x}_{j}$, and $\mathbf{K}_{ij}$ is the so-called kernel matrix \cite{sohrab2018subspace}.

The hyperparameters of the S-SVDD models are fine-tuned using the validation set and we subsequently evaluated their performance on the test sets, where all samples that did not belong to the target class were treated as outliers. We choose the values of the hyperparameters from the following ranges by performing 5-fold cross-validation over the training set: 
\begin{itemize}
  \item$\beta\in\{10^{-2},10^{-1},10^{0},10^{1},10^{2}\},$
  \item$C\in\{0.1,0.2,0.3,0.4,0.5\}$,
  \item$\sigma\in\{10^{-1},10^{0},10^{1},10^{2},10^{3}\}$,
  \item$d\in\{1,2,3,4,5,10,20\}$,
  \item$\eta\in\{10^{-1},10^{0},10^{1},10^{2},10^{3}\}$,
\end{itemize}

where the maximum number of iterations is fixed to 10 for all variants of S-SVDD.

We used geometric mean (GM) as our performance metric for finding the best-performing hyper-parameters. The GM is calculated by taking the square root of the product of the true positive rate (TPR) and the true negative rate (TNR). The TPR represents the proportion of correctly classified positive samples, while the TNR represents the proportion of correctly classified negative samples.

The advantage of using the GM metric is that it takes into account both the TPR and TNR, and thus it reflects the model's ability to detect both positive and negative samples accurately. Additionally, it is less sensitive to changes in the dataset's class distribution than other performance metrics, such as accuracy or F1 score. Therefore, it is a reliable choice as a performance metric when the goal is to balance the detection of both positive and negative samples since the datasets are significantly imbalanced.

\psubsection{Results}

Tables \ref{tab:res} and \ref{tab:res2} present the obtained results over Salinas-A and Indian Pines datasets, respectively. The performance of S-SVDD classifiers with different regularizations $\psi_{0-3}$ and mappings are evaluated for two datasets. The performance of different variants of S-SVDD is measured by considering each class separately as a target class and the rest all together as an outliers class. The evaluation metric is in the range of 0 to 1, with higher values indicating better performance. Accordingly, the results show that different regularization terms have consistent results in the linear case, while more varying results are noticed in the non-linear case for different regularization terms. Overall, the results suggest that the choice of regularization should be tailored to the specific target class in S-SVDD, especially while using the non-linear data description.

\begin{table*}[h!]
\caption{GM results over Salinas-A dataset using Subspace Support Vector Data Description (S-SVDD) with different regularizations $\psi_{0-3}$ and mappings.}
\label{tab:res}
\vspace{0.2cm}
\centering
\resizebox{.55\linewidth}{!}{
\begin{tabular}{cccccccc}
\toprule
 & \multicolumn{1}{c}{\multirow{2}{*}{\textbf{Method}}} & \multicolumn{6}{c}{\textbf{Target Class}} \\ \cmidrule{3-8}

& \multicolumn{1}{c}{} & \multicolumn{1}{c}{\textbf{1}} & \multicolumn{1}{c}{\textbf{2}} & \multicolumn{1}{c}{\textbf{3}} & \multicolumn{1}{c}{\textbf{4}} & \multicolumn{1}{c}{\textbf{5}} & \multicolumn{1}{c}{\textbf{6}} \\ \midrule

\multicolumn{1}{c}{\vspace{0.05cm}\multirow{4}{*}{\rotatebox[origin=c]{90}{\textit{\textbf{Linear}}}}}\hspace{-0.3cm} & S-SVDD $\psi_0$ & $0.996$ & $0.736$ & $0.862$ & $0.996$ & $0.999$ & $0.952$ \\

\multicolumn{1}{c}{\vspace{0.05cm}} & S-SVDD $\psi_1$ & $0.996$ & $0.731$ & $0.863$ & $0.995$ & $0.999$ & $0.953$ \\

\multicolumn{1}{c}{\vspace{0.05cm}} & S-SVDD $\psi_2$ & $0.996$ & $0.730$ & $0.863$ & $0.996$ & $0.999$ & $0.952$ \\

\multicolumn{1}{c}{\vspace{0.05cm}} & S-SVDD $\psi_3$ & $0.996$ & $0.732$ & $0.863$ & $0.995$ & $0.999$ & $0.952$ \\ \midrule

\multicolumn{1}{c}{\vspace{0.05cm}\multirow{4}{*}{\rotatebox[origin=c]{90}{\textit{\textbf{Non-linear}}}}}\hspace{-0.3cm} & S-SVDD $\psi_0$ & $0.894$ & $0.746$ & $0.865$ & $0.994$ & $0.999$ & $0.937$ \\

\multicolumn{1}{c}{\vspace{0.05cm}} & S-SVDD $\psi_1$ & $0.644$ & $0.764$ & $0.865$ & $0.994$ & $0.999$ & $0.952$ \\

\multicolumn{1}{c}{\vspace{0.05cm}} & S-SVDD $\psi_2$ & $0.792$ & $0.626$ & $0.824$ & $0.994$ & $0.999$ & $0.913$ \\

\multicolumn{1}{c}{\vspace{0.05cm}} & S-SVDD $\psi_3$ & $0.984$ & $0.574$ & $0.866$ & $0.992$ & $0.999$ & $0.907$ \\ \bottomrule
\end{tabular}}
\end{table*}

\begin{table*}[h!]
\caption{GM results over Indian Pines dataset using Subspace Support Vector Data Description (S-SVDD) with different regularizations $\psi_{0-3}$ and mappings.}
\label{tab:res2}
\vspace{0.2cm}
\centering
\resizebox{1\linewidth}{!}{
\begin{tabular}{cccccccccccccccccc}
\toprule
 & \multicolumn{1}{c}{\multirow{2}{*}{\textbf{Method}}} & \multicolumn{16}{c}{\textbf{Target Class}} \\ \cmidrule{3-18}

& \multicolumn{1}{c}{} & \multicolumn{1}{c}{\textbf{1}} & \multicolumn{1}{c}{\textbf{2}} & \multicolumn{1}{c}{\textbf{3}} & \multicolumn{1}{c}{\textbf{4}} & \multicolumn{1}{c}{\textbf{5}} & \multicolumn{1}{c}{\textbf{6}} & \multicolumn{1}{c}{\textbf{7}} & \multicolumn{1}{c}{\textbf{8}} & \multicolumn{1}{c}{\textbf{9}} & \multicolumn{1}{c}{\textbf{10}} & \multicolumn{1}{c}{\textbf{11}} & \multicolumn{1}{c}{\textbf{12}} & \multicolumn{1}{c}{\textbf{13}} & \multicolumn{1}{c}{\textbf{14}} & \multicolumn{1}{c}{\textbf{15}} & \multicolumn{1}{c}{\textbf{16}}\\ \midrule

\multicolumn{1}{c}{\vspace{0.05cm}\multirow{4}{*}{\rotatebox[origin=c]{90}{\textit{\textbf{Linear}}}}}\hspace{-0.3cm} & S-SVDD $\psi_0$ & $0.982$ & $0.576$ & $0.584$ & $0.529$ & $0.833$ & $0.962$ & $0.986$ & $0.986$ & $0.964$ & $0.701$ & $0.566$ & $0.571$ & $0.987$ & $0.944$ & $0.903$ & $0.770$ \\

\multicolumn{1}{c}{\vspace{0.05cm}} & S-SVDD $\psi_1$ & $0.978$ & $0.577$ & $0.585$ & $0.527$ & $0.833$ & $0.962$ & $0.971$ & $0.985$ & $0.892$ & $0.702$ & $0.571$ & $0.569$ & $0.989$ & $0.941$ & $0.904$ & $0.783$ \\

\multicolumn{1}{c}{\vspace{0.05cm}} & S-SVDD $\psi_2$ & $0.982$ & $0.612$ & $0.585$ & $0.530$ & $0.834$ & $0.962$ & $0.977$ & $0.985$ & $0.892$ & $0.701$ & $0.569$ & $0.569$ & $0.988$ & $0.940$ & $0.903$ & $0.798$ \\

\multicolumn{1}{c}{\vspace{0.05cm}} & S-SVDD $\psi_3$ & $0.968$ & $0.576$ & $0.582$ & $0.529$ & $0.833$ & $0.962$ & $0.979$ & $0.987$ & $0.892$ & $0.70$ & $0.566$ & $0.571$ & $0.988$ & $0.944$ & $0.903$ & $0.757$ \\ \midrule

\multicolumn{1}{c}{\vspace{0.05cm}\multirow{4}{*}{\rotatebox[origin=c]{90}{\textit{\textbf{Non-linear}}}}}\hspace{-0.3cm} & S-SVDD $\psi_0$ & $0.944$ & $0.611$ & $0.637$ & $0.678$ & $0.827$ & $0.966$ & $0.827$ & $0.985$ & $0.852$ & $0.676$ & $0.597$ & $0.481$ & $0.975$ & $0.907$ & $0.888$ & $0.736$ \\

\multicolumn{1}{c}{\vspace{0.05cm}} & S-SVDD $\psi_1$ & $0.982$ & $0.661$ & $0.604$ & $0.541$ & $0.825$ & $0.962$ & $0.868$ & $0.985$ & $0.562$ & $0.620$ & $0.565$ & $0.331$ & $0.975$ & $0.939$ & $0.893$ & $0.873$ \\

\multicolumn{1}{c}{\vspace{0.05cm}} & S-SVDD $\psi_2$ & $0.950$ & $0.612$ & $0.641$ & $0.670$ & $0.830$ & $0.966$ & $0.867$ & $0.985$ & $0.738$ & $0.686$ & $0.432$ & $0.589$ & $0.973$ & $0.943$ & $0.882$ & $0.862$ \\

\multicolumn{1}{c}{\vspace{0.05cm}} & S-SVDD $\psi_3$ & $0.952$ & $0.605$ & $0.643$ & $0.711$ & $0.829$ & $0.966$ & $0.763$ & $0.985$ & $0.843$ & $0.676$ & $0.224$ & $0.608$ & $0.976$ & $0.907$ & $0.881$ & 0.873 \\ \bottomrule
\end{tabular}}
\end{table*}

\psection{Conclusion}
In conclusion, the high-dimensional nature and imbalanced classes of hyperspectral images pose challenges for traditional machine learning algorithms. One-class classifiers are useful in cases where the training data is from a single class only, but they still face challenges in handling the curse of dimensionality. To address these challenges, we leverage S-SVDD for one-class classification of hyperspectral images. Our experiments on two benchmark HSI datasets show that the proposed approach can effectively tackle the curse of dimensionality and the imbalanced nature of HSI data. In the future, we first aim to focus on embedding the graph information \cite{sohrab2023graph} in the optimization process of the proposed approach. We also consider utilizing neighbouring pixels' spectral features for a future study, since they are usually correlated and give better scattering information, allowing enhanced performance.
\ack
This work was supported by the NSF-Business Finland project AMALIA. Foundation for Economic Education (Grant number: 220363) funded the work of Fahad Sohrab at Haltian.
\bibliography{refs}
\bibliographystyle{unsrt}

\end{paper}

\end{document}